\documentclass[letterpaper, 10 pt, conference]{ieeeconf}  %

\IEEEoverridecommandlockouts                              %

\usepackage{hyperref} 
\usepackage{mathrsfs}
\usepackage{amsfonts}
\usepackage{amsmath}
\usepackage[flushleft]{threeparttable}
\usepackage{tablefootnote}
\usepackage{multirow}
\usepackage{graphicx}
\usepackage{subfigure}
\let\labelindent\relax
\usepackage{enumitem}
\usepackage{hanging}
\usepackage{color}
\usepackage{tikz}
\usetikzlibrary{calc,arrows,decorations.markings}
\usepackage{balance}
\usepackage{tabularx}
\usepackage{pifont}
\usepackage{amssymb}
\newcommand{\cmark}{\ding{51}}
\newcommand{\xmark}{\ding{55}}

\definecolor{darkgreen}{RGB}{0,127,0}
\definecolor{darkred}{RGB}{200,0,0}

\newcommand{\R}{\mathbb{R}}  %
\newcommand{\calL}{\ensuremath{{\cal L}}}

\graphicspath{{image}}

\usepackage{booktabs}  %

\title{\LARGE \bf
Single-Stage Keypoint-Based Category-Level \\ Object Pose Estimation from an RGB Image}

\author{Yunzhi Lin$^{1,2}$, Jonathan Tremblay$^{1}$, Stephen Tyree$^{1}$, Patricio A.  Vela$^{2}$, 
Stan Birchfield$^{1}$ 
\\ $^{1}$NVIDIA: {\tt\small \{jtremblay, styree, sbirchfield\}@nvidia.com}
\\ $^{2}$Georgia Institute of Technology: {\tt\small \{yunzhi.lin, pvela\}@gatech.edu}
\thanks{Work was completed while the first author was an intern at NVIDIA.}%
\thanks{This work was supported in part by NSF Award \#2026611.}%
\thanks{Project:  \url{https://sites.google.com/view/centerpose}}%
}

\begin{document}

\maketitle
\thispagestyle{empty}
\pagestyle{empty}

\begin{abstract}

Prior work on 6-DoF object pose estimation has largely focused on \emph{instance-level} processing,
in which a textured CAD model is available for each object being detected.
\emph{Category-level} 6-DoF pose estimation represents an important step toward developing robotic vision systems that operate in unstructured, real-world scenarios.
In this work, we propose a single-stage, keypoint-based approach for category-level object pose estimation that operates on unknown object instances within a known category using a single RGB image as input.
The proposed network performs 2D object detection, detects 2D keypoints, estimates 6-DoF pose, and regresses relative bounding cuboid dimensions.
These quantities are estimated in a sequential fashion, leveraging the recent idea of convGRU for propagating information from easier tasks to those that are more difficult. 
We favor simplicity in our design choices:  generic cuboid vertex coordinates, single-stage network, and monocular RGB input.
We conduct extensive experiments on the challenging Objectron benchmark, outperforming state-of-the-art methods on the 3D IoU metric (27.6\% higher than the MobilePose single-stage approach and 7.1\% higher than the related two-stage approach).

\end{abstract}

\section{Introduction}

Scene awareness is a fundamental skill for robotic manipulators to operate in unconstrained environments.
This ability includes locating objects and their poses, 
also known as the 6-DoF pose estimation problem (\emph{i.e.}, 6 degrees of freedom, from 3D position + orientation).
Accurate, real-time pose information of nearby objects in the scene would allow robots to engage in semantic interaction.

The problem of pose estimation is a rich topic in the computer vision community, yet most existing methods have focused on \textit{instance-level} object pose estimation
\cite{xiang2018posecnn,tremblay2018deep,wang2019densefusion}. 
Such methods suffer from lack of scalability: a detector trained for the `cracker box' in the YCB object dataset~\cite{calli2015ram:ycb}, for example,
will work reliably on instances of that specific object (similar size and texture),
 but the same network may fail to detect an instance with different textures (\emph{e.g.}, due to seasonal promotional changes) and will yield poor pose estimates of boxes with matching texture but different size.
Further, the detector is expected to \textit{ignore} all other types of cracker boxes or food-containing cuboids.
As a result, the number of \textit{instance-level} detectors required increases rapidly with scene complexity.

\begin{figure}[t]
  \centering
  \includegraphics[width=\linewidth]{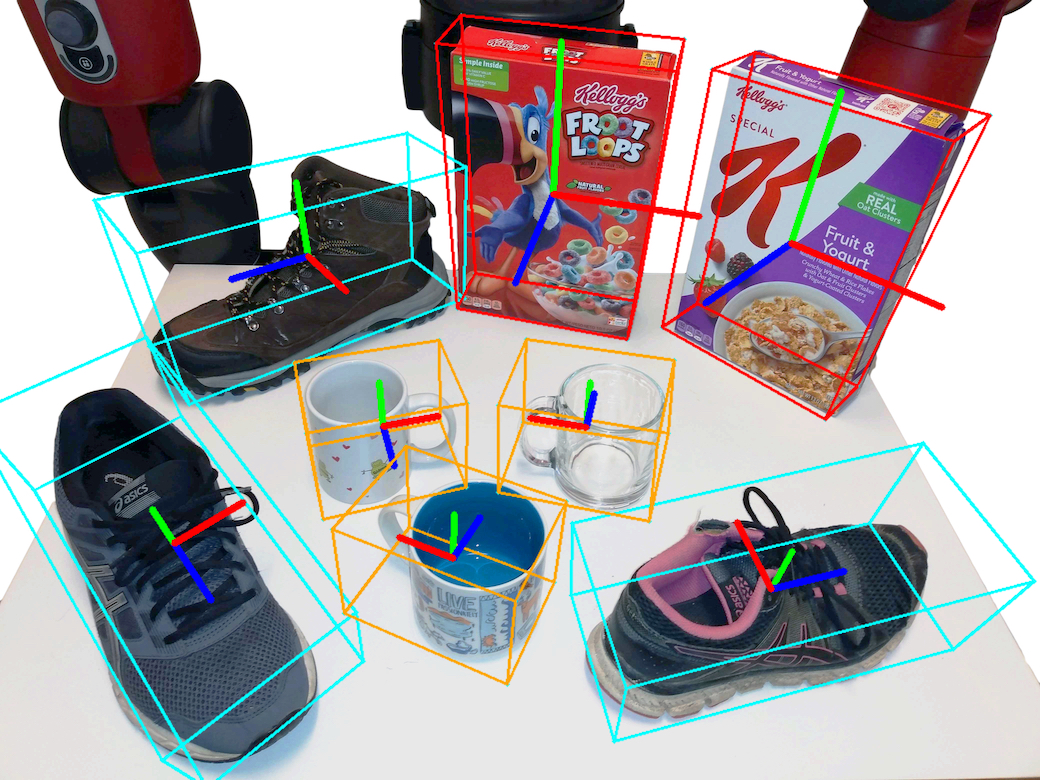}  
  \caption{  Given a single RGB image containing previously unseen instances of known categories (in this case \textit{cereal boxes}, \textit{cups}, and \textit{shoes}), our proposed method detects objects and estimates 6-DoF poses and 3D bounding box dimensions up to a scale factor.  We use a separate network for each category.}
  \label{fig:abstract}
  \vspace*{-4 ex}
\end{figure}

To alleviate this challenge, we focus on \textit{category-level} pose estimation.
Our goal is to detect and infer the pose and relative size of all objects within a specific category using a monocular RGB image processed by a single-stage neural network.
For example, in Fig.~\ref{fig:abstract} our network predicts the pose of each object in a specific category, along with its dimensions (expressed as relative width, height, and length), using a single set of trained weights.

A few recent works have considered \textit{category-level} object pose estimation~\cite{pavlakos2017,wang2019normalized,ke2020gsnet,tian2020shape,chen2020learning,chen2020category,manhardt2020cps++}.
By removing the requirement of exact CAD models of object instances \emph{at inference time}, these methods promise to scale better for real-world applications.
To train the network, one could use a large collection of 3D CAD models (\emph{e.g.}, from ShapeNet~\cite{shapenet2015}) to render synthetic samples with complex annotations, such as pixel-wise segmentation masks or normalized object coordinate spaces (NOCS~\cite{wang2019normalized}). Yet the domain gap between synthetic and real data  remains an obstacle, sometimes even after fine-tuning on annotated real-world data \cite{kortylewski2018training}. In the meantime, many techniques require depth in addition to color (RGB) images \cite{wang2019normalized, tian2020shape, chen2020learning}. While monocular RGB-based methods~\cite{hou2020mobilepose,ahmadyan2021objectron} have not received much attention, they have great potential for wide applicability and for handling certain material properties, such as transparent or dark surfaces, that are difficult for depth sensors.

In this work, we address the aforementioned challenges and limitations by proposing a simple and efficient RGB-based approach (without depth) that only requires oriented 3D bounding box annotations at training time, and thus does not require CAD models for training. This design decision allows us to take advantage of large collections of real-world images, such as the Objectron dataset \cite{ahmadyan2021objectron}, which are annotated with \textit{category-level} 3D bounding boxes.

Inspired by %
CenterNet~\cite{zhou2019objects}, we use a single-stage neural network to regress object locations in the image, 2D keypoint projections of 3D bounding box vertices, and relative dimensions of the bounding box. The simple design of generic 3D bounding box keypoint regression allows the same method to be applied to a wide variety of categories.
To better handle %
intra-class shape variability, we adopt a two-fold representation of both displacements and heatmaps for keypoint detection. This choice achieves a good balance between accuracy and design complexity, as shown in our experiments. 
Furthermore, the single-stage design of our network avoids the complexity of multi-stage networks~\cite{ahmadyan2021objectron} and enables end-to-end learning, as well as potentially faster training time.

To improve the tractability of regressing so many outputs, the network output modalities are grouped by increasing difficulty, and we use a
convolutional gated recurrent unit (convGRU)~\cite{ballas2016delving, gao2020monocular} to compute each output group from an underlying sequentially-refined hidden state. 
This way, the difficulty of predicting the later groups is alleviated by using information stored in the hidden state from previous groups. 
Once objects have been detected in image space, our approach of estimating relative cuboid dimensions allows us to leverage robust off-the-shelf P$n$P algorithms for pose estimation.

Our work makes the following contributions:
\begin{itemize}
    \item A single-stage keypoint-based network for detecting previously unseen objects from known categories and estimating their 6-DoF poses and relative bounding box dimensions from a monocular RGB input.
    \item 
    Demonstration of the benefit of directly predicting relative dimensions of the 3D bounding cuboid for category-level pose estimation, as well as the benefit of sequential feature association to improve the accuracy of estimating scale information for difficult cases.
    \item Experiments showing that the proposed method achieves state-of-the-art performance on the large-scale Objectron dataset \cite{ahmadyan2021objectron}. 
\end{itemize}

\section{Related Work}
{\bf Instance-level object pose estimation.} 
Assuming that a 3D (possibly textured) CAD model is available for each object class at both training and inference time, these methods aim to infer each object's position and orientation in 3D.
Current approaches can be divided into two types: template matching and regression. Template matching techniques align known 3D CAD models to the observed 3D point clouds \cite{zeng2017multi}, 2D images \cite{li2018deepim,sundermeyer2018implicit} or local descriptors \cite{choi20123d,birdal2015point}. State-of-the-art template-based methods have demonstrated impressive results on public benchmarks like BOP \cite{hodan2018bop}. %

Regression-based methods directly regress the 6-DoF pose \cite{xiang2018posecnn} or predict the image coordinates of 2D projected keypoints to establish 2D-3D correspondences for solving the 6-DoF pose using a P$n$P algorithm \cite{rad2017bb8,tekin2018real,oberweger2018making,tremblay2018deep,hou2020mobilepose}. Other works have explored different ways to better represent objects, including dense coordinate maps \cite{xiang2018posecnn}, keypoints \cite{peng2019pvnet}, and symmetry correspondences \cite{song2020hybridpose}. Although our method is inspired by keypoint regression techniques, we do not require 3D CAD models. As a result, the dimensions of the object have to be estimated in addition to pose.%

{\bf Category-level object pose estimation.} Recently, researchers have begun to explore category-level object pose estimation, which does not require instance-specific 3D object models at test time. Wang et al.~\cite{wang2019normalized} propose a normalized object coordinate space (NOCS) to serve as a common reference frame for 6-DoF pose and size estimation of unseen objects. Their proposed network is based on two-stage Mask R-CNN \cite{he2017mask}, which predicts the NOCS map for a pose fitting algorithm that accepts the depth map as input. However, 3D meshes were still required during training to calculate the NOCS map, requiring a synthetic training dataset.%

Subsequent RGBD works mainly focus on fusing RGB and depth information.  Chen et al.~\cite{chen2020learning} propose a correspondence-free approach by learning a canonical shape space for input RGBD images.  Their approach also eases network training by matching pose-dependent and pose-independent features separately. Tian et al.~\cite{tian2020shape} model the deformation from the categorical shape prior to the object model by latent embeddings, then recover 6-DoF pose by estimating a similarity transformation between observed points and NOCS map.

To the best of our knowledge, only a few approaches attempt category-level pose estimation from monocular RGB images.
Manhardt et al.~\cite{manhardt2020cps++} propose to regress shape and pose parameters and recover depth, while Chen et al.~\cite{chen2020category} propose a neural analysis-by-synthesis approach. However, both of these methods still require synthetic CAD models (e.g., ShapeNet~\cite{shapenet2015}) at training time.
Hou et al.~\cite{hou2020mobilepose} present a single-stage light-weight model with two heads regressing to the centroid location and the 3D bounding box keypoints, respectively, from an RGB image. Similarly, Ahmadyan et al.~\cite{ahmadyan2021objectron} introduce a two-stage architecture for 3D bounding box keypoint regression from an RGB image.
Both approaches are trained directly on real images from Objectron and thus do not require CAD models or synthetic data.\footnote{Both methods have reported impressive real-time performance on a mobile GPU (36~fps for~\cite{hou2020mobilepose} and 83~fps for~\cite{ahmadyan2021objectron}). Their networks have been heavily optimized, which is beyond the scope of our work.}
These methods do not take the object dimensions into account when solving for pose. They instead directly lift the 2D predicted keypoints to 3D via a modified EP$n$P algorithm \cite{lepetit2009epnp} by fixing the homogeneous barycentric coordinates.
In contrast, as we show experimentally, our approach achieves better performance by directly regressing the relative dimensions of the cuboid and using an off-the-shelf P$n$P algorithm.

\begin{figure*}[t]
  \centering
    \includegraphics[width=\textwidth]{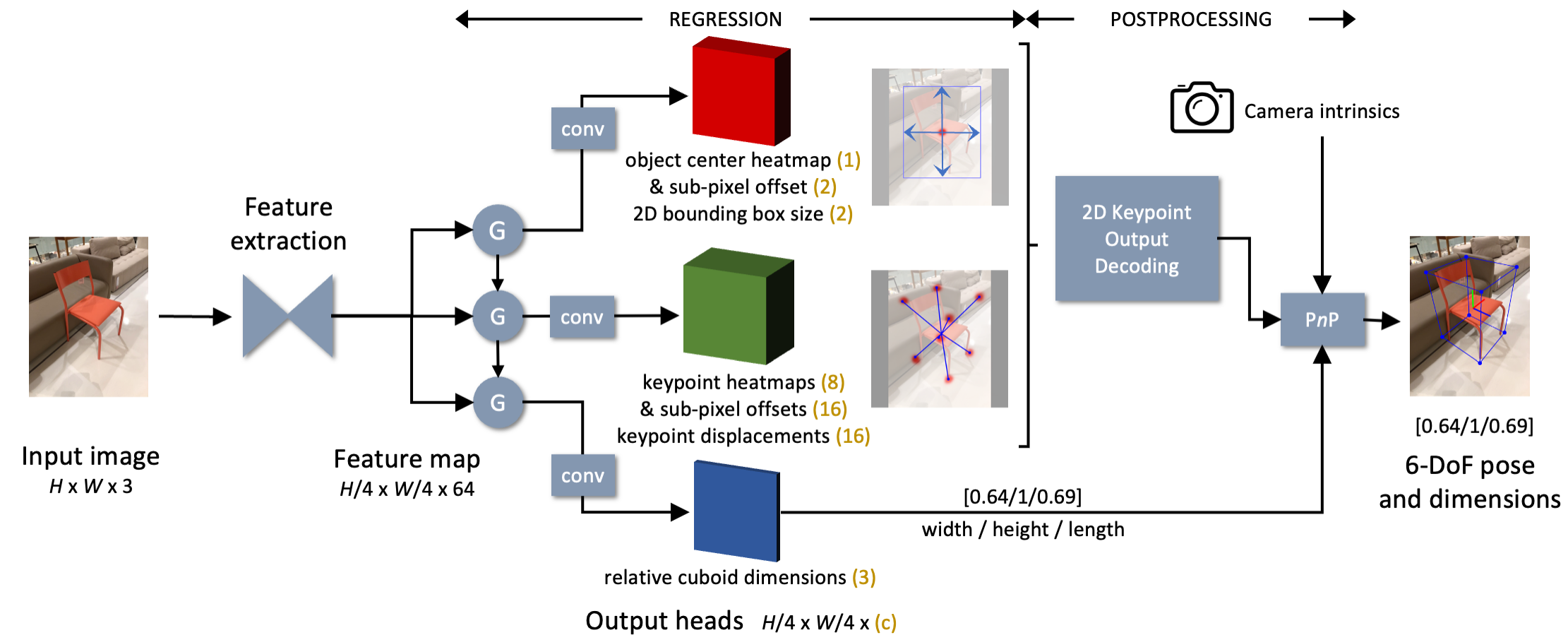}
  \caption{Overview of our method. For an input image $I \in \R^{H \times W \times 3}$, the backbone network extracts a feature map $\Phi(I) \in \R^{H/4 \times W/4 \times 64}$. A convGRU module associates features into three sequential groups, regressing a total of seven different outputs.  Each output is in $\R^{H/4 \times W/4 \times c}$, where the number in parentheses $(c)$ is the number of channels. The corresponding modalities are extracted from the feature map via lookup at the detected center.  %
  Finally, we use the decoded 2D keypoints, known camera intrinsics, and the dimensions of the 3D bounding box to obtain a final 6-DoF pose (up to a scale factor) via P$n$P.
    \label{fig:pipeline}}
  \vspace*{-3 ex}
\end{figure*}

\section{Approach}

Our approach to category-level object pose estimation is illustrated in Figure~\ref{fig:pipeline}.
We follow the lead of prior correspondence-based methods~\cite{rad2017bb8,tekin2018real,oberweger2018making,tremblay2018deep,hou2020mobilepose} by predicting 2D image projections of the corners of the 3D bounding cuboid, followed by P$n$P to compute pose.
Inspired by the recent success of works based on the CenterNet architecture \cite{zhou2019objects,gao2020monocular,liu2020smoke,wang2020centermask}, we employ a single-stage network to make all predictions, including the relative dimensions of the cuboid, which are necessary for P$n$P. To alleviate the difficulty of inferring 3D structure information from a 2D input, we propose to use a convGRU module~\cite{gao2020monocular} to predict outputs grouped in increasing order of difficulty. Further details are provided below.

\subsection{Architecture Design}
\label{approach: arch}
The network takes an RGB image of resolution $H \times W \times 3$, with source images re-scaled and padded as needed so that $W=H=512$.
We adopt DLA-34~\cite{yu2018deep} combined with upsampling as the backbone network, where
hierarchical aggregation connections are augmented by deformable convolutional layers \cite{zhu2019deformable}. 
The backbone network produces multiple intermediate feature maps of spatial resolutions ranging from $H/4 \times W/4$ to $H/32 \times W/32$, which are aggregated in a single $H/4 \times W/4 \times 64$ output.

The network has a total of seven output heads arranged in three groups. 
For each output head, a $3 \times 3$ convolutional layer with 256 channels followed by a $1 \times 1$ convolution layer is used to process the output of the corresponding convGRU module. %
Outputs are predicted as dense heatmaps or regression maps, but are accessed sparsely in correspondence with detected object centers, as described subsequently.

{\bf Object detection branch.}
The primary output of the entire network, at least conceptually, is the \emph{object center heatmap} whose peaks indicate the centers of the 2D bounding boxes for detected objects.\footnote{We considered defining the object center as the projection of the 3D bounding box center, as in \cite{gao2020monocular,liu2020smoke}, 
but obtained much better results using the center of the 2D bounding box.}
Other output maps are accessed \textit{w.r.t.} the object center: If a peak is found in the object center heatmap at location $(c_x,c_y)$, the values at $(c_x,c_y)$ in the remaining outputs are associated with this object. %
To recover the discretization error resulting from the heatmap output resolution, we regress a local 2D \emph{object center sub-pixel offset} map following \cite{zhou2019objects}.%

Since the Objectron dataset \cite{ahmadyan2021objectron} does not provide 2D bounding box annotations, we define it as the smallest axis-aligned rectangle that encloses the extreme points %
of the projected ground truth 3D bounding box.

{\bf Keypoint detection branch.}
Our network uses two methods to predict the 2D coordinates of 3D bounding box vertices projected into image space.
First, we regress 2D keypoint \emph{displacement} vectors from the bounding box center point.
Second, we output a set of 8 \emph{keypoint heatmaps} whose peaks indicate the 2D coordinates of the projected 3D vertices. These peaks are not accessed at the coordinates of the object's center like other outputs. (Further details are given in Section~\ref{sec:decode}.)
Training labels for keypoint heatmaps are generated by a Gaussian kernel centered at the ground truth keypoint coordinates with variance determined by the size of the 2D bounding box.
As above, to mitigate discretization error, we also output a local 2D \emph{keypoint sub-pixel offset} for each vertex.

{\bf Cuboid dimensions branch.} Since category-level pose estimation assumes that we do not have access to the CAD model of the target object instance, we use a final output branch to estimate the \emph{relative dimensions} (width, height, 
length) of the 3D bounding cuboid.
\emph{Relative} values are predicted to avoid the need to implicitly estimate absolute depth from a monocular RGB image, which is a fundamentally ill-posed problem.
(For example, we do not know whether we are viewing a full-size actual chair or a toy chair.)
Relative values also allow us to apply our network to images obtained with different camera intrinsics without having to retrain the network.
Since many target objects in daily life have a canonical orientation when resting on the ground, we choose the up $(y)$ axis as the primary axis. Ground truth scale labels are considered to be $(x/y,1,z/y)$, with the ratios $x/y$ and $z/y$ estimated by the network. 
Unlike 3D vehicle detection approaches~\cite{gao2020monocular,liu2020smoke} that use an exponential offset between 3D
dimensions and the category-specific dimension
template, we directly regress each ratio since the objects we encounter exhibit much more diversity in aspect ratios than are found in vehicles.

\subsection{convGRU Feature Association}

We expect that some network outputs are more difficult to learn than the others.
Heuristically, we divided them into the three groups discussed previously, as shown in Figure~\ref{fig:pipeline}:
1) \textit{object center heatmap,  object center sub-pixel offset, and 2D bounding box size}; 
2)  \textit{x-y displacements to keypoint, keypoint heatmaps, and keypoint sub-pixel offsets}; and
3) \textit{relative cuboid dimensions}. The last group is the most difficult to estimate since 3D structure has to be implicitly deduced from 2D appearance. 
We hypothesize that keypoints are more easily found once the object centroid and 2D bounding box are estimated, and similarly that bounding box dimensions are more easily predicted after keypoints have been found.

Inspired by Gao et al.~\cite{gao2020monocular}, this grouping strategy and sequential output construction is naturally formulated by assigning different output groups to different ``timesteps'' in a recurrent neural network.\footnote{Timesteps here simply refer to recurrent iterations; there is no temporal aspect to the input data.} Given an input image $I$, the $i^\text{th}$ output ($i=1,\ldots,7$) is represented as:
\begin{equation}
y_{i}=\Psi_{i}\left(G_{t}\left(\Phi(I), h_{t-1}\right)\right), %
\end{equation}
where $\Phi(I)$ denotes the feature map from the backbone network, $G_{t}(\cdot)$ represents the GRU at timestep $t$, $h_{t-1}=G_{t-1}\left(\Phi(I), h_{t-2}\right)$ denotes the hidden state produced by the GRU cell at the previous timestep, $h_0=0$, and $\Psi_{i}$ is the fully convolutional network for the $i^\text{th}$ output. 
The timesteps $t=1,2,3$ correspond to the three output groups used in our method.

We adopt a single-layer convolutional GRU network where all the convolution layers in the convGRU are set with stride $ = 1$, kernel size $ = 3$, and output channels $ = 64$. The output from a later timestep will have access to the hidden states flowing from the previous timestep, which implements the idea of output grouping and sequential feature association.

\subsection{2D Keypoint Output Decoding \label{sec:decode}} 

The outputs of the network are decoded and assembled in the following manner. 
First, a $3 \times 3$ max pooling operation is applied on the heatmap of the 2D object center, which serves as an efficient alternative to non-maximum suppression~\cite{zhou2019objects}.
For each detected center point, displacement-based keypoint locations are then given by the 2D x-y displacements under the center point. 
Next, heatmap-based keypoint locations are extracted by finding high confidence peaks in the corresponding heatmaps that are within a margin of the 2D object bounding box.
Both estimates of keypoint locations are adjusted according to the sub-pixel offsets, then input, along with the estimated relative cuboid dimensions, to the Levenberg-Marquardt version of P$n$P~\cite{abdel2015direct}.

\subsection{Loss Function}

{\bf Focal loss.}
We employ penalty-reduced focal losses \cite{lin2017focal} $\mathcal{L}_{\text {p}_{cen}}$ and $\mathcal{L}_{\text {p}_{key}}$ in a point-wise manner for the center point and keypoint heatmaps, respectively:
\begin{equation} 
\mathcal{L}_\text{p}=\frac{-1}{N} \sum_{i j}\left\{\begin{array}{ll}
(1-\hat{Y}_{i j})^{\alpha} \log (\hat{Y}_{i j}) & \text{if}\ Y_{i j}=1 \\
(1-Y_{i j})^{\beta}(\hat{Y}_{i j})^{\alpha} \log(1-\hat{Y}_{i j}) &\text{otherwise}
\end{array}\right.
\end{equation}

\noindent where $\hat{Y}_{i, j}$ is the predicted score at the heatmap location $(i, j)$ and $Y_{i, j}$ represents the ground-truth value of each point assigned by Gaussian kernel. $N$ is the number of center points in the image, $\alpha$ and $\beta$ are the hyper-parameters of the focal loss, which are set to $\alpha=2, \beta=4$, following~\cite{zhou2019objects}.

{\bf L1 loss.} 
The center sub-pixel offset loss, $\calL_\text{off}$, is computed using an L1 loss.
Let $\hat{O}$ represent the predicted offset, $p$ the ground truth center point, and $R$ the output stride, then the low-resolution equivalent of $p$ is $\tilde{p}=\left\lfloor\frac{p}{R}\right\rfloor.$ The sub-pixel offset loss is:
\begin{equation}
\mathcal{L}_\text{off}=\frac{1}{N} \sum_{p}\left\|\hat{O}_{\tilde{p}}-\left(\frac{p}{R}-\tilde{p}\right)\right\|.
\end{equation}
The keypoint sub-pixel offset loss, $\calL_\text{offkey}$, is computed similarly. The 2D bounding box size, $\calL_\text{bbox}$, the keypoint displacement loss, $\calL_\text{dis}$, and the relative cuboid dimensions loss, $\calL_\text{dim}$, are also computed using an L1 loss w.r.t. their label values.

{\bf Overall loss.} 
The overall training objective is the weighted combination of seven
loss terms:
\begin{equation}
\begin{aligned}
\mathcal{L}_{\text {all }}=& \lambda_{\text {p}_{cen}} \mathcal{L}_{\text {p}_{cen}}+\lambda_{\text {off }} \mathcal{L}_{\text {off }}+\lambda_{\text {bbox }} \mathcal{L}_{\text {bbox }} \\
&+\lambda_{\text {p}_{key}} \mathcal{L}_{\text {p}_{key}}+\lambda_{\text {offkey }} \mathcal{L}_{\text {offkey }} \\
&+\lambda_{\text {dis }} \mathcal{L}_{\text {dis }}+\lambda_{\text {dim }} \mathcal{L}_{\text {dim }}, 
\end{aligned}
\end{equation}
where $
\lambda_{\text {p}_{cen}} 
=\lambda_{\text{off }} 
=\lambda_{\text {p}_{key}} 
=\lambda_\text {offkey } 
=\lambda_{\text {dis }} 
=\lambda_\text{dim} =1$, and $\lambda_{\text {bbox }}=0.1$.

\subsection{Implementation Details} 

The network was trained with a batch-size of 32 on 4 NVIDIA V-100 GPUs for 140 epochs, starting with pretrained weights from ImageNet.
Data augmentation included random flip, scaling, cropping, and color jittering. 
We chose Adam as the optimizer with an initial learning rate of {2.5e-4}, dropping 10x at both 90 and 120 epochs.
An average of 36 hours was required to train
one category (using between 8k to 32k training images depending on the category). Inference speed is around 15~fps on a NVIDIA GTX 1080Ti GPU.

\section{Experimental Results \label{ExpResults}}

\subsection{Dataset}

The Objectron dataset~\cite{ahmadyan2021objectron} is a newly proposed benchmark for monocular RGB category-level 6-DoF object pose estimation. 
The dataset consists of 15k annotated video clips with over 4M annotated frames.
Objects are from the following nine categories: bikes, books, bottles, cameras, cereal boxes, chairs, cups, laptops, and shoes.
Each object is annotated with a 3D bounding cuboid, which describes the object's position and orientation with respect to the camera, as well as the cuboid dimensions. 
For each video recording, the camera moves around a stationary object, capturing it from different angles. Additional metadata includes camera poses, sparse point clouds, and surface planes, with the latter assuming that the object rests on the ground plane, which yields an absolute scale factor.
For training, we extract frames by temporally downsampling the original videos at 15~fps. %
For testing, we evaluate all the test samples in each category from the official release of the dataset for straightforward comparison with other methods.

The cup category contains both cups and mugs, where the former do not have handles.
Therefore, we manually differentiate these by training a separate network for each. %
We also noticed ambiguities in mug instances where the handle is not consistently oriented.
To solve this problem, we manually checked all videos and rotated some of the ground truth bounding boxes by 180 degrees to ensure consistent orientation. The cup/mug split will be released along with our code.

For symmetric objects like cups, we follow the idea of Wang et al.~\cite{wang2019normalized} to generate multiple ground truth labels $\left\{{\mathbf{y}}_{1}, \ldots, {\mathbf{y}}_{|\theta|}\right\}$ during the training phase, rotating $|\theta|$=12 times around the symmetry axis. %
The symmetric loss is then computed as $\calL_\text{sym}=\min _{i=1, \ldots,|\theta|} \calL\left({\mathbf{y}}_{i}, \hat{\mathbf{y}}\right),$ where $\hat{\mathbf{y}}$ denotes the prediction, and $\calL$ is the asymmetric loss.

\subsection{Metrics}

Following the Objectron dataset~\cite{ahmadyan2021objectron}, we adopt the average precision (AP) of 3D IoU metric proposed by \cite{hou2020mobilepose} with a threshold of 50$\%$ to evaluate 3D detection and object dimension estimation.
The 2D pixel projection error metric computes the mean normalized distance between the projections of 3D bounding box keypoints given
the estimated and ground truth pose.
For viewpoint estimation, we report the AP of azimuth and elevation with a threshold of $15^{\circ}$ and $10^{\circ}$, respectively.
For symmetric object categories (bottle$^*$ and cup$^*$), we rotate
the estimated bounding box along the symmetry axis N times ($N=100$ following \cite{ahmadyan2021objectron}) and
evaluate the prediction w.r.t. each rotated instance. The reported number is the instance that maximizes 3D IoU or minimizes 2D pixel projection error, respectively. 
Although the cup category also includes mug instances which are asymmetric, we still treat them as symmetric for a fair comparison with \cite{ahmadyan2021objectron}.
For the comparison on relative dimension prediction, we use mean relative dimension error, which computes the relative error of the relative dimension across all predictions $\frac{1}{n}\sum_{i=1}^{n} \frac{\left|\hat{y}_{i}-y_{i}\right|}{y_{i}}$, where $\hat{y}_i$ denotes the prediction and $y_i$ denotes the ground truth.

\begin{figure*}[t]
  \scalebox{0.87}{
  \begin{tikzpicture}[inner sep=0pt, outer sep=0pt]
     \node[anchor=south west] (Bike_gt) at (0in,0in)
      {\includegraphics[height=1.10in,clip=true,trim=0.25in 0.25in 0in 0.25in]{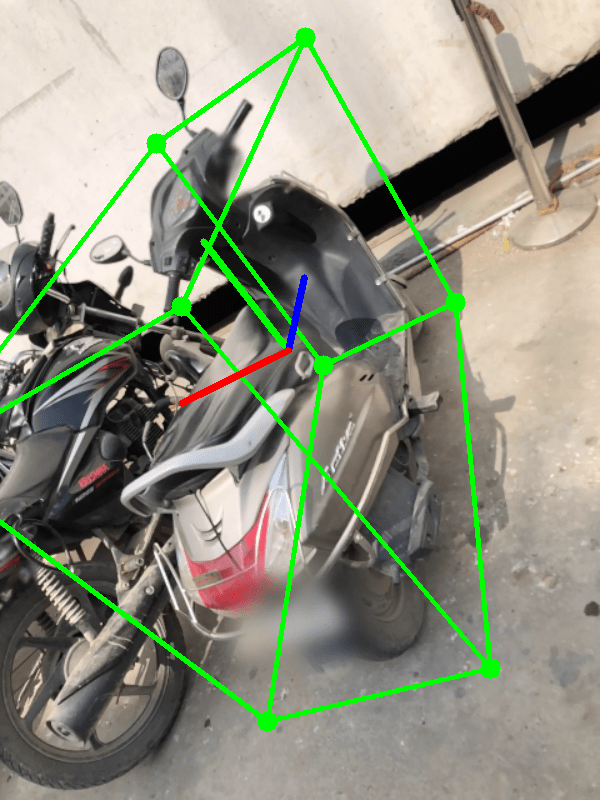}};
     \node[anchor=south,xshift=0pt,yshift=-10pt] at (Bike_gt.south)
     {\small [0.55/1/1.34]};
     \node[anchor=north west,yshift=-15pt] (Bike_pred) at (Bike_gt.south west)
      {\includegraphics[height=1.10in,clip=true,trim=0.25in 0.25in 0in 0.25in]{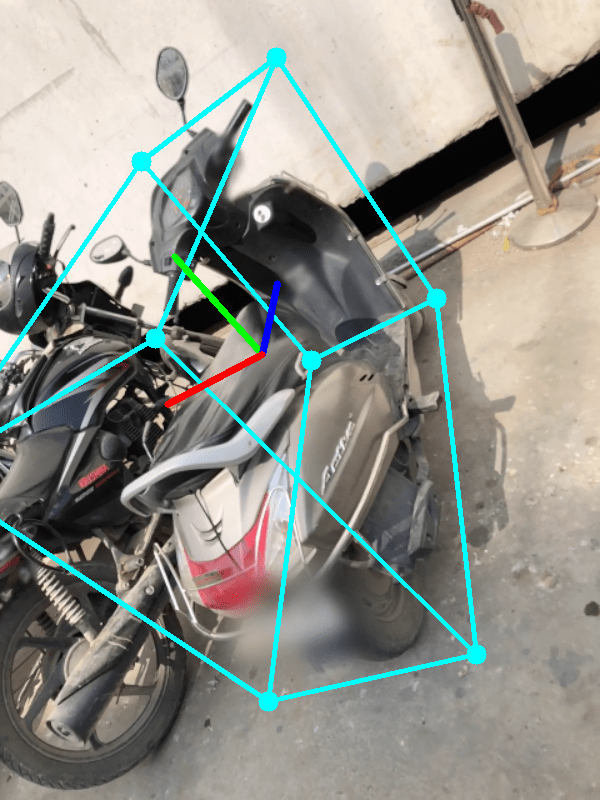}};
    \node[anchor=south,xshift=0pt,yshift=-10pt] at (Bike_pred.south)
    {\small [0.57/1/1.39]};
     \node[anchor=south,xshift=0pt,yshift=-27pt] at (Bike_pred.south)
    {\small a) Bike};
    \node[anchor=base,rotate=90,yshift=7pt] at (Bike_gt.west)
      {\small 6D pose + size};
    \node[anchor=base,rotate=90,yshift=17pt] at (Bike_gt.west)
      {\small Ground truth};
    \node[anchor=base,rotate=90,yshift=7pt] at (Bike_pred.west)
      {\small 6D pose + size};
    \node[anchor=base,rotate=90,yshift=17pt] at (Bike_pred.west)
      {\small Predicted};
      
     \node[anchor=south west, xshift=2pt] (Book_gt) at (Bike_gt.south east)
      {\includegraphics[height=1.10in,clip=true,trim=0in 0in 0in 0in]{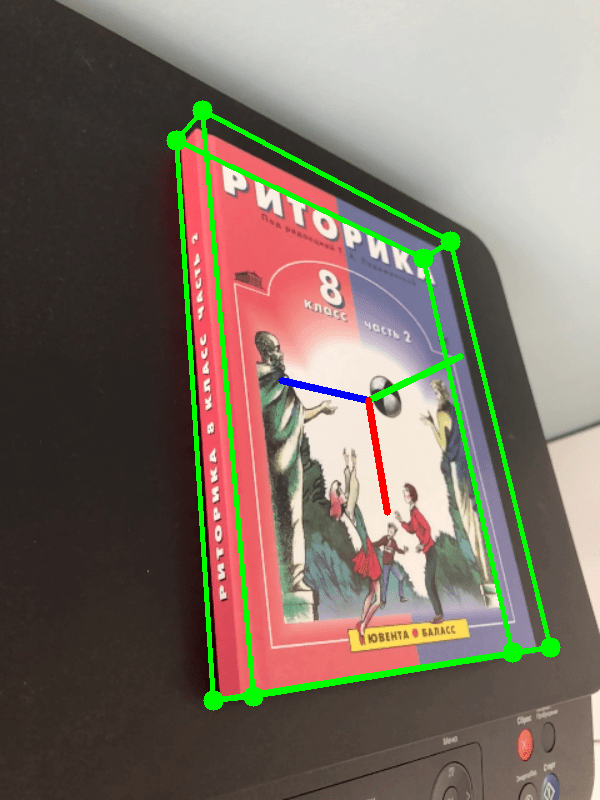}};
     \node[anchor=south,xshift=0pt,yshift=-10pt] at (Book_gt.south)
     {\small [11.00/1/8.50]};
     \node[anchor=north west,yshift=-15pt] (Book_pred) at (Book_gt.south west)
      {\includegraphics[height=1.10in,clip=true,trim=0in 0in 0in 0in]{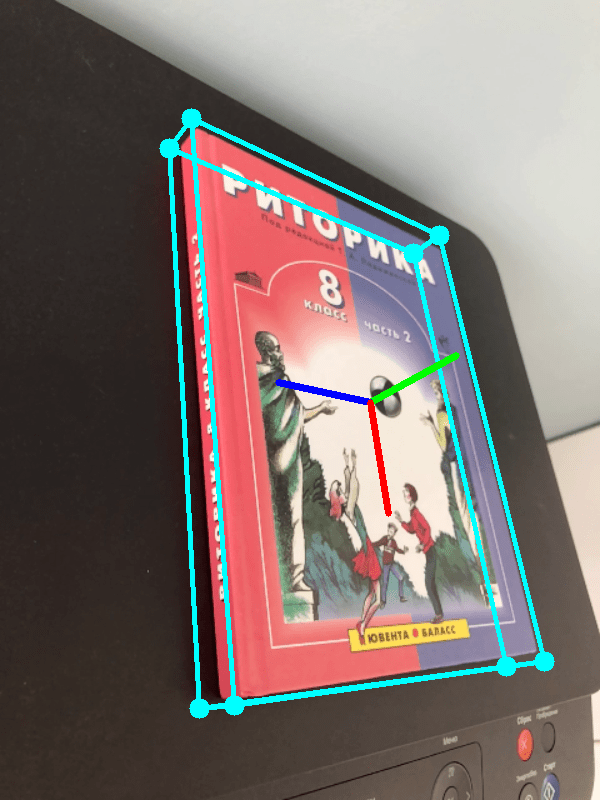}}; 
  \node[anchor=south,xshift=0pt,yshift=-10pt] at (Book_pred.south)
    {\small [11.54/1/8.22]};
     \node[anchor=south,xshift=0pt,yshift=-27pt] at (Book_pred.south)
    {\small b) Book};
    
     \node[anchor=south west, xshift=2pt] (Bottle_gt) at (Book_gt.south east)
      {\includegraphics[height=1.10in,clip=true,trim=0in 0in 0in 0in]{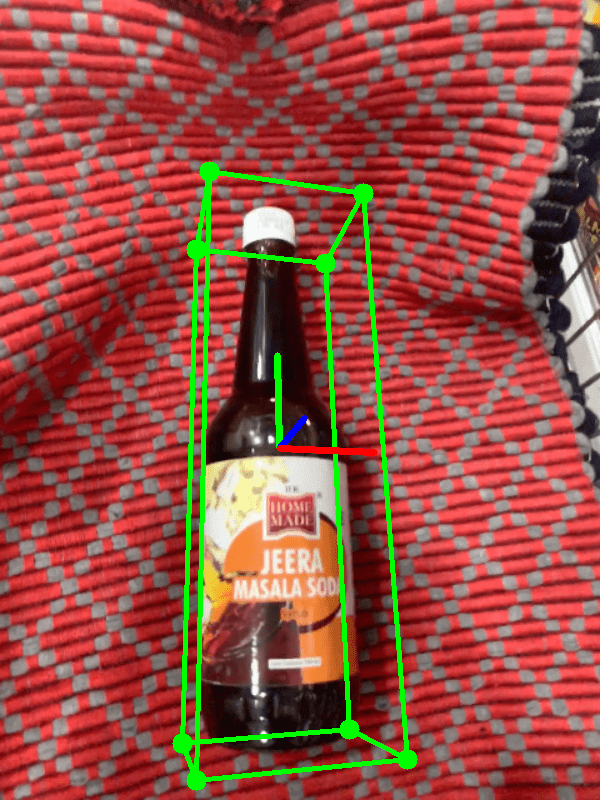}};
     \node[anchor=south,xshift=0pt,yshift=-10pt] at (Bottle_gt.south)
     {\small [0.31/1/0.27]};
     \node[anchor=north west,yshift=-15pt] (Bottle_pred) at (Bottle_gt.south west)
      {\includegraphics[height=1.10in,clip=true,trim=0in 0in 0in 0in]{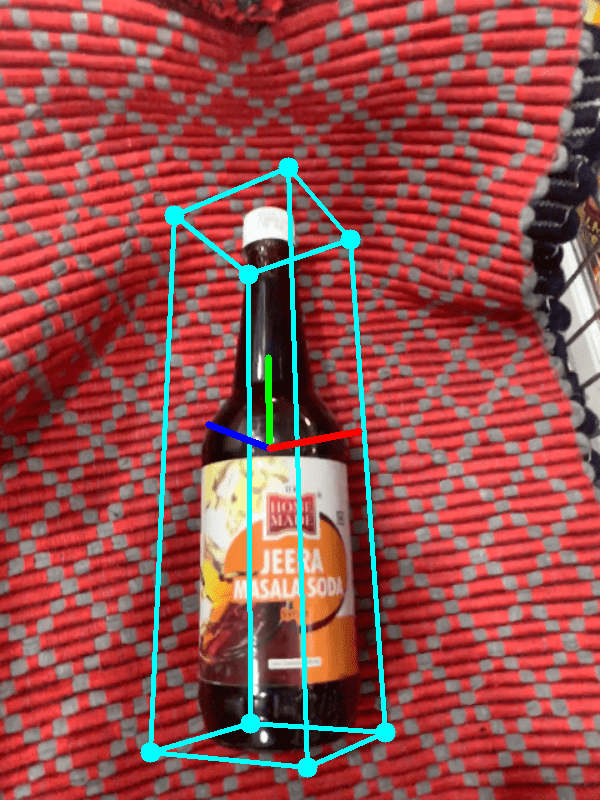}};
        \node[anchor=south,xshift=0pt,yshift=-10pt] at (Bottle_pred.south)
    {\small [0.28/1/0.28]};
     \node[anchor=south,xshift=0pt,yshift=-27pt] at (Bottle_pred.south)
    {\small c) Bottle$^{*}$};
    
     \node[anchor=south west, xshift=2pt] (Camera_gt) at (Bottle_gt.south east)
      {\includegraphics[height=1.10in,clip=true,trim=0in 0in 0in 0in]{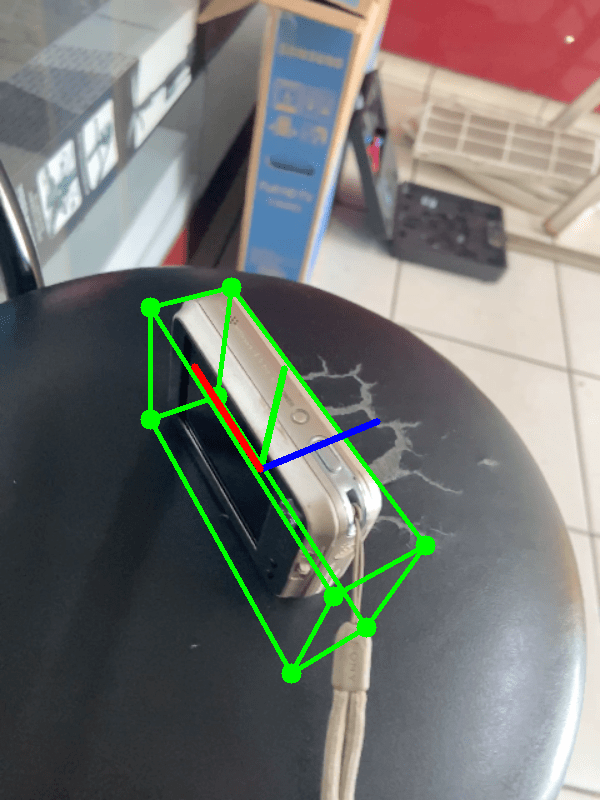}};
     \node[anchor=south,xshift=0pt,yshift=-10pt] at (Camera_gt.south)
     {\small [1.83/1/0.50]};
     \node[anchor=north west,yshift=-15pt] (Camera_pred) at (Camera_gt.south west)
      {\includegraphics[height=1.10in,clip=true,trim=0in 0in 0in 0in]{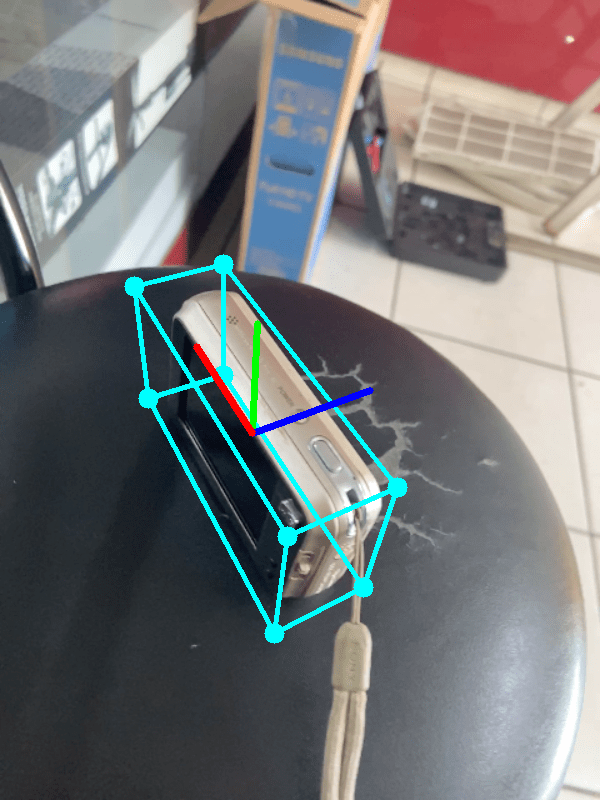}};
         \node[anchor=south,xshift=0pt,yshift=-10pt] at (Camera_pred.south)
    {\small [1.93/1/0.59]};
     \node[anchor=south,xshift=0pt,yshift=-27pt] at (Camera_pred.south)
    {\small d) Camera};
    
     \node[anchor=south west, xshift=2pt] (Cereal_box_gt) at (Camera_gt.south east)
      {\includegraphics[height=1.10in,clip=true,trim=0in 0in 0in 0in]{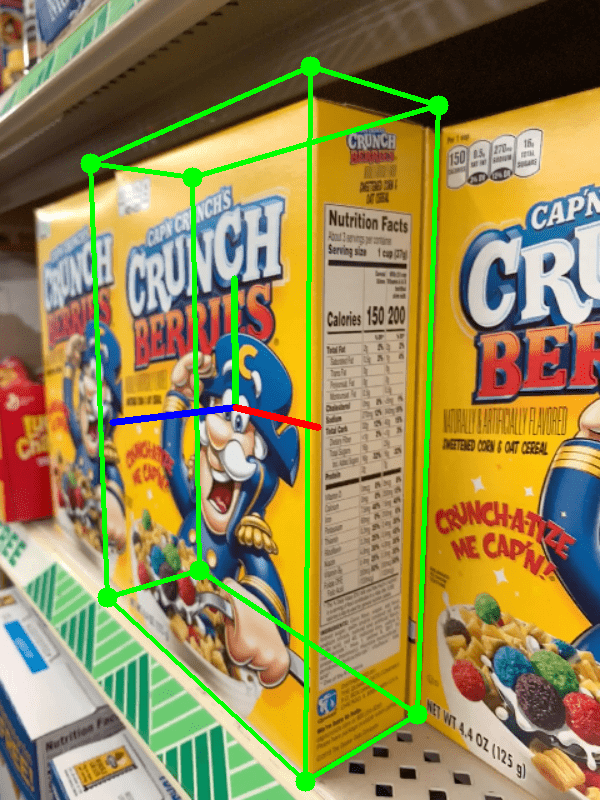}};
     \node[anchor=south,xshift=0pt,yshift=-10pt] at (Cereal_box_gt.south)
     {\small [0.75/1/0.25]};
     \node[anchor=north west,yshift=-15pt] (Cereal_box_pred) at (Cereal_box_gt.south west)
      {\includegraphics[height=1.10in,clip=true,trim=0in 0in 0in 0in]{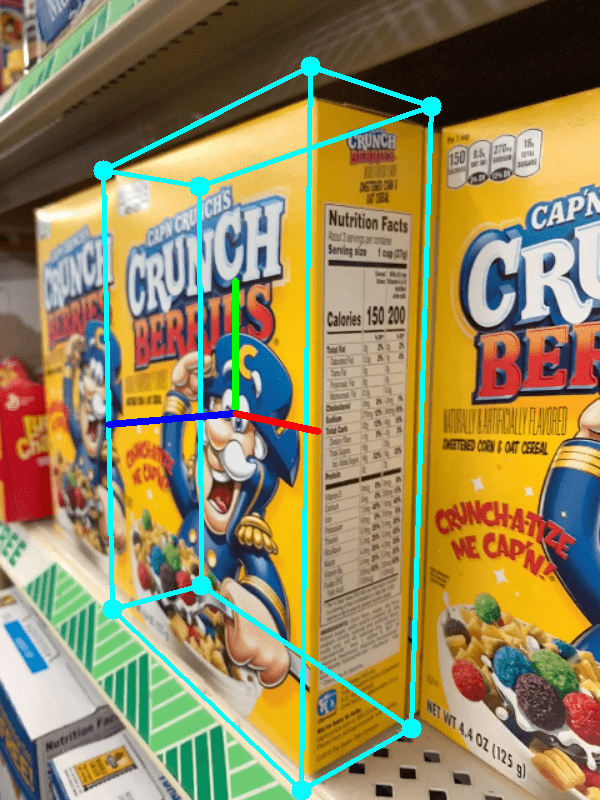}};
    \node[anchor=south,xshift=0pt,yshift=-10pt] at (Cereal_box_pred.south)
    {\small [0.72/1/0.23]};
     \node[anchor=south,xshift=0pt,yshift=-27pt] at (Cereal_box_pred.south)
    {\small e) Cereal\_box};
    
     \node[anchor=south west, xshift=2pt] (Chair_gt) at (Cereal_box_gt.south east)
      {\includegraphics[height=1.10in,clip=true,trim=0in 0in 0in 0in]{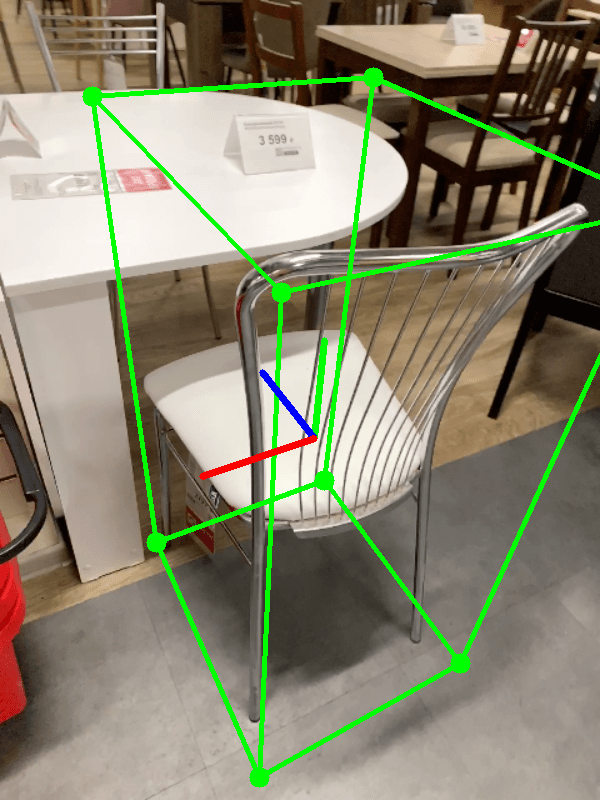}};
     \node[anchor=south,xshift=0pt,yshift=-10pt] at (Chair_gt.south)
     {\small [0.47/1/0.57]};
     \node[anchor=north west,yshift=-15pt] (Chair_pred) at (Chair_gt.south west)
      {\includegraphics[height=1.10in,clip=true,trim=0in 0in 0in 0in]{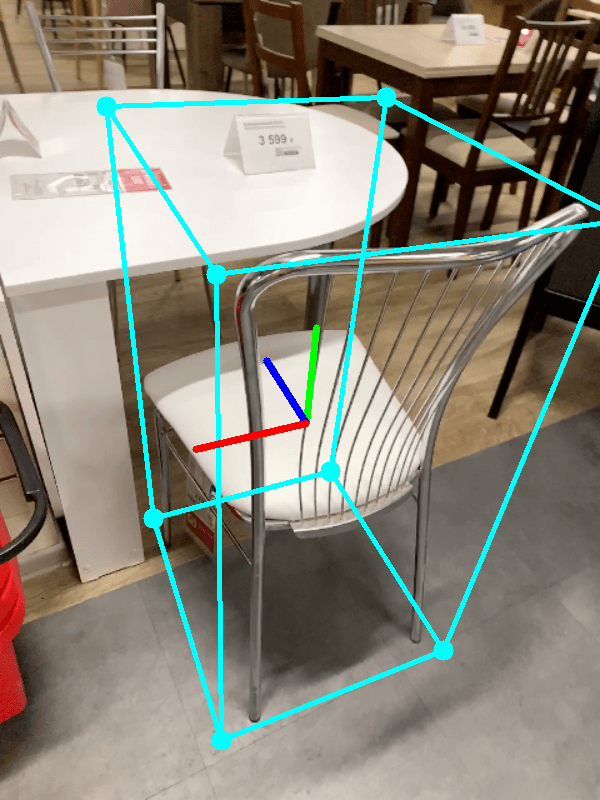}};
        \node[anchor=south,xshift=0pt,yshift=-10pt] at (Chair_pred.south)
    {\small [0.51/1/0.58]};
     \node[anchor=south,xshift=0pt,yshift=-27pt] at (Chair_pred.south)
    {\small f) Chair};
    
     \node[anchor=south west, xshift=2pt] (Cup_gt) at (Chair_gt.south east)
      {\includegraphics[height=1.10in,clip=true,trim=0in 0in 0in 0in]{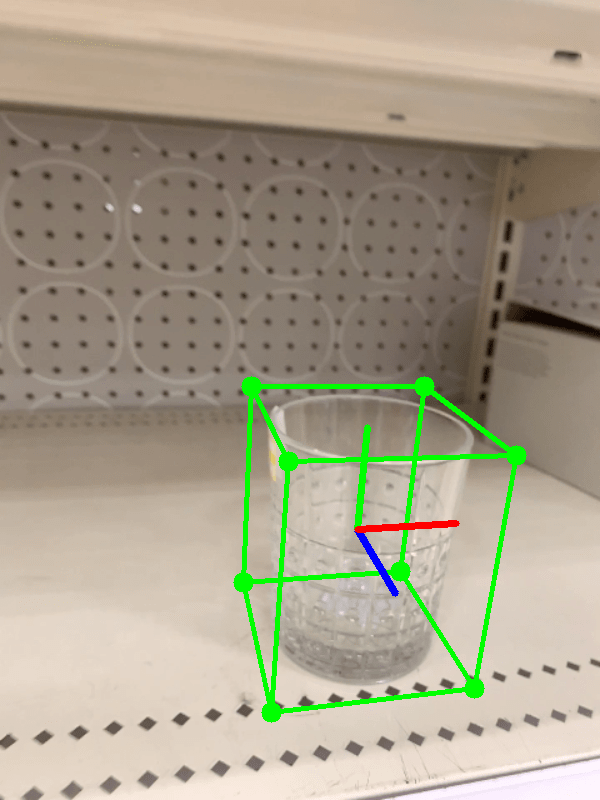}};
     \node[anchor=south,xshift=0pt,yshift=-10pt] at (Cup_gt.south)
     {\small [0.82/1/0.82]};
     \node[anchor=north west,yshift=-15pt] (Cup_pred) at (Cup_gt.south west)
      {\includegraphics[height=1.10in,clip=true,trim=0in 0in 0in 0in]{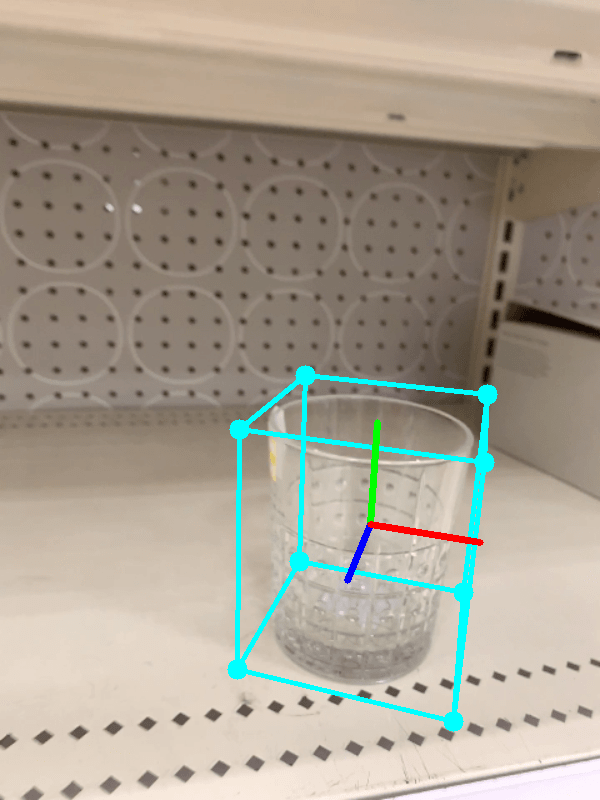}};
        \node[anchor=south,xshift=0pt,yshift=-10pt] at (Cup_pred.south)
    {\small [0.83/1/0.85]};
     \node[anchor=south,xshift=0pt,yshift=-27pt] at (Cup_pred.south)
    {\small g) Cup$^{*}$};
    
     \node[anchor=south west, xshift=2pt] (Laptop_gt) at (Cup_gt.south east)
      {\includegraphics[height=1.10in,clip=true,trim=0in 0in 0in 0in]{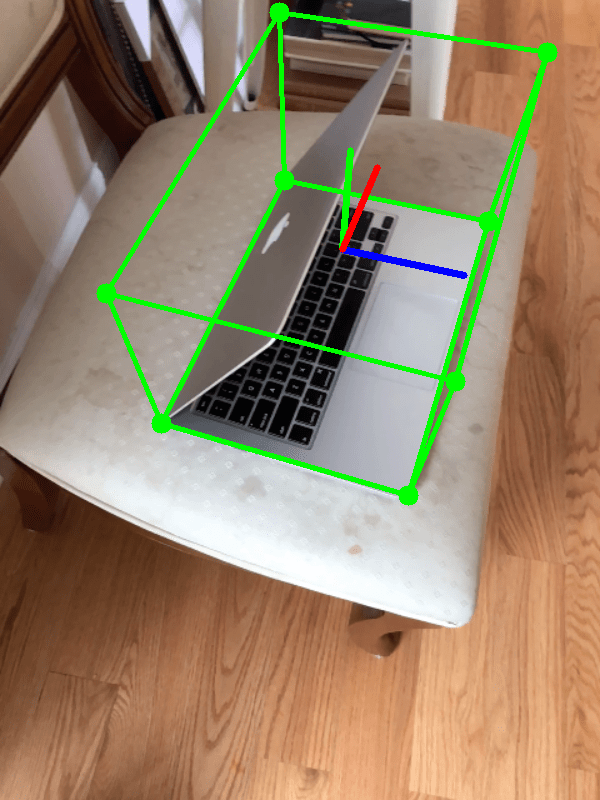}};
     \node[anchor=south,xshift=0pt,yshift=-10pt] at (Laptop_gt.south)
     {\small [1.67/1/1.19]};
     \node[anchor=north west,yshift=-15pt] (Laptop_pred) at (Laptop_gt.south west)
      {\includegraphics[height=1.10in,clip=true,trim=0in 0in 0in 0in]{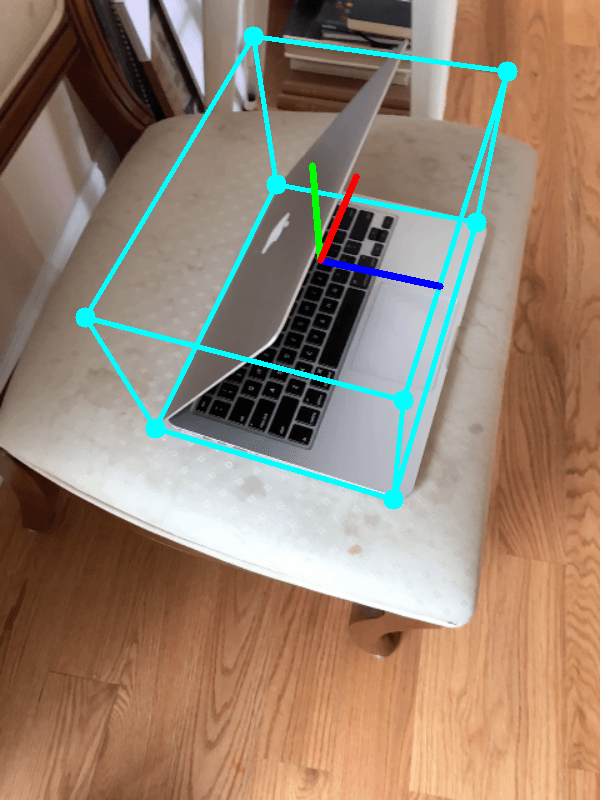}};
    \node[anchor=south,xshift=0pt,yshift=-10pt] at (Laptop_pred.south)
    {\small [1.84/1/1.29]};
     \node[anchor=south,xshift=0pt,yshift=-27pt] at (Laptop_pred.south)
    {\small h) Laptop};
    
     \node[anchor=south west, xshift=2pt] (Shoe_gt) at (Laptop_gt.south east)
      {\includegraphics[height=1.10in,clip=true,trim=0in 0in 0in 0in]{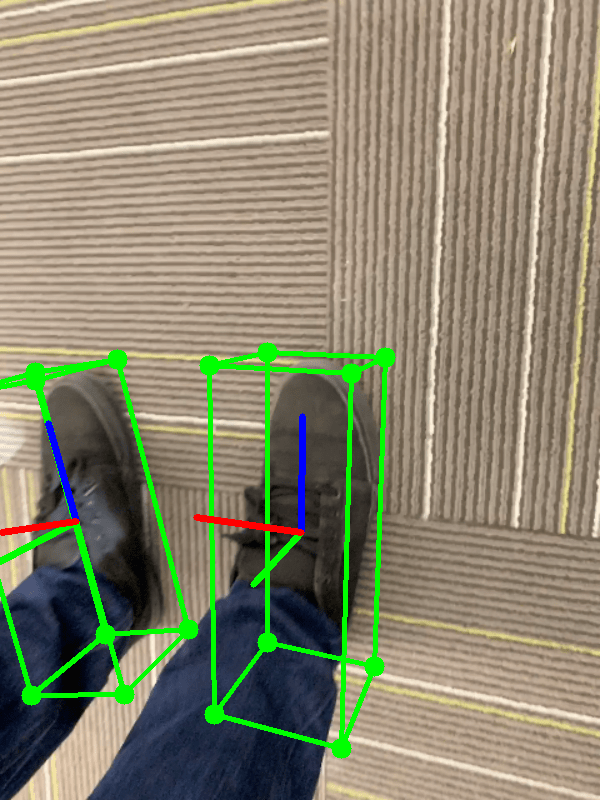}};
     \node[anchor=south,xshift=0pt,yshift=-10pt] at (Shoe_gt.south)
     {\small [0.96/1/2.57]};
     \node[anchor=north west,yshift=-15pt] (Shoe_pred) at (Shoe_gt.south west)
      {\includegraphics[height=1.10in,clip=true,trim=0in 0in 0in 0in]{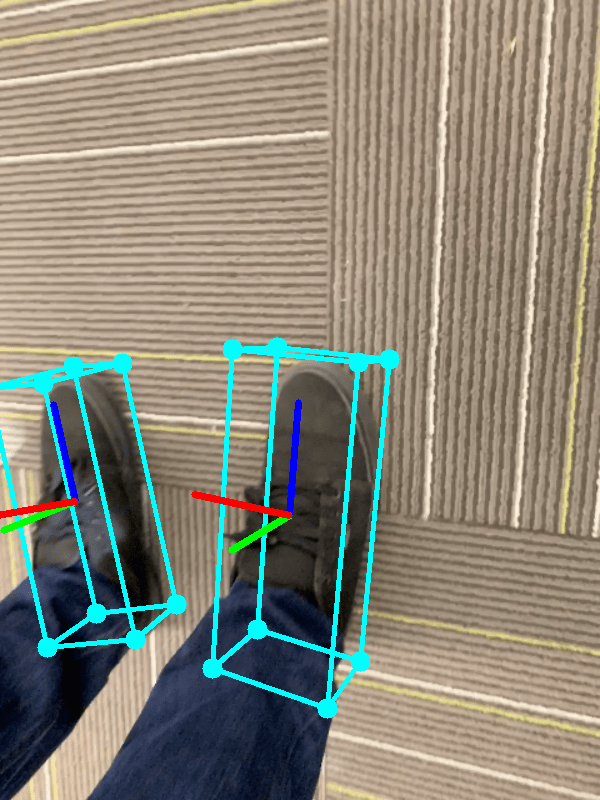}};
    \node[anchor=south,xshift=0pt,yshift=-10pt] at (Shoe_pred.south)
    {\small [1.36/1/3.49]};
    \node[anchor=south,xshift=0pt,yshift=-18pt] at (Shoe_pred.south)
    {\small [1.28/1/3.27]};
     \node[anchor=south,xshift=0pt,yshift=-27pt] at (Shoe_pred.south)
    {\small i) Shoe};
    
  \end{tikzpicture}}
  \caption{Sample results of our method on the Objectron dataset \cite{ahmadyan2021objectron}, where the number below represents the relative dimensions of the 3D bounding box. The proposed method handles large intra-class shape variance, diverse viewpoints, noisy backgrounds, and transparent surfaces.
     \label{fig:exp_qualitative}}
  \vspace*{-2 ex}
\end{figure*}
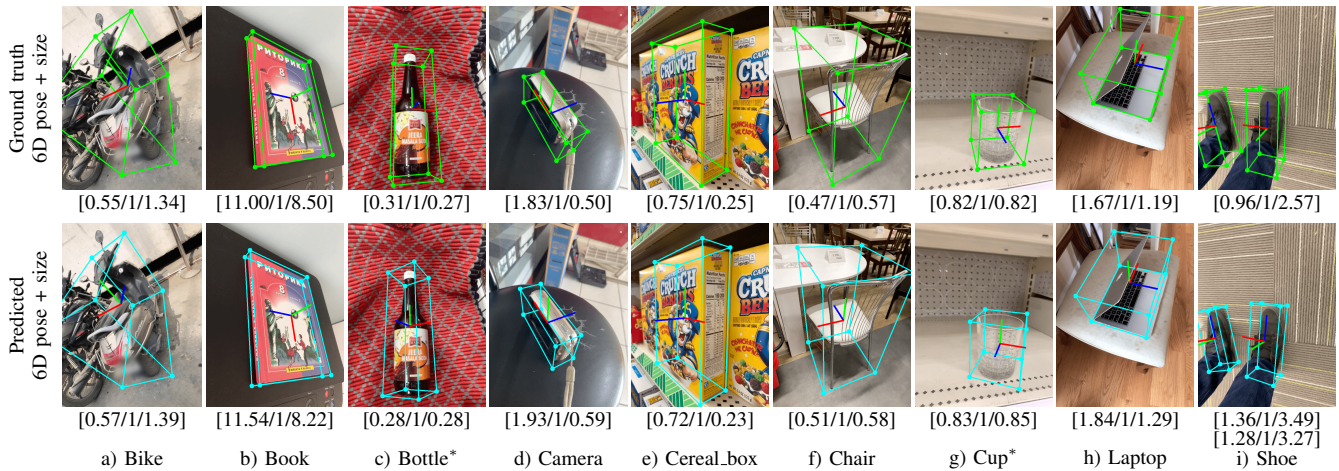

\subsection{Category-Level 6-DoF Pose and Size Estimation}

We compare our proposed method with two state-of-the-art methods: single-stage MobilePose \cite{hou2020mobilepose} and a two-stage network \cite{ahmadyan2021objectron}.
To the best of our knowledge, these are the only methods available for the Objectron dataset.
Results for 
3D IoU, 2D pixel projection error, 
and AP for azimuth and elevation are shown in Table~\ref{tab:comparison_combined}. 
We also present qualitative results in Figure~\ref{fig:exp_qualitative}.
Our method significantly outperforms MobilePose on all metrics, while the two-stage method \cite{ahmadyan2021objectron} achieves better performance on the metric of mean pixel error of 2D projection but falls behind on 3D IoU metric. Their two-stage structure allows the keypoint detector to operate at a higher image resolution for better keypoint location performance, but also limits its ability for end-to-end training and fast scale-up to more categories (since the two networks have to be trained independently). Moreover, they do not take  the  object  dimensions into account but rather rely on a modified EP$n$P algorithm by fixing homogeneous barycentric coordinates across all the cases, which leads to an unstable solution of the 2D--3D correspondence equation \cite{lepetit2009epnp}.

\begin{table*}
 \vspace*{0.06in}
\caption{Pose estimation comparison on the Objectron test set~\cite{ahmadyan2021objectron}.}
\label{tab:comparison_combined}
  \centering
\begin{tabular}{cccccccccccc}
\toprule
      Stage & Method     & Bike   & Book   & Bottle$^{*}$ & Camera & Cereal\_box & Chair  & Cup$^{*}$ & Laptop & Shoe   & Mean   \\ 
\midrule
\multicolumn{12}{c}{Average precision at 0.5 3D IoU ($\uparrow$)} \\
\cmidrule(r){5-7}
One & MobilePose \cite{hou2020mobilepose} & 0.3109 & 0.1797 & 0.5433 & 0.4483 & 0.5419 & 0.6847 & 0.3665 & 0.5225 & 0.4171 & 0.4461 \\
Two &Two-stage \cite{ahmadyan2021objectron} & 0.6127 & 0.5218          & 0.5744                           & \textbf{0.8016} & 0.6272          & \textbf{0.8505} & 0.5388                        & 0.6735 & 0.6606 & 0.6512 \\ 
One  & Ours                                                     & \textbf{0.6419} & \textbf{0.5565} & \textbf{0.8021} & 0.7188 & \textbf{0.8211} & 0.8471 & \textbf{0.7704} & \textbf{0.6766} & \textbf{0.6618} & \textbf{0.7218} \\
\\
\multicolumn{12}{c}{Mean pixel error of 2D projection of cuboid vertices ($\downarrow$)} \\
\cmidrule(r){5-7}
One & MobilePose \cite{hou2020mobilepose} & 0.1581 & 0.0840 & 0.0818 & 0.0773 & 0.0454 & 0.0892 & 0.2263 & 0.0736 & 0.0655 & 0.1001\\
Two  & Two-stage  \cite{ahmadyan2021objectron}  & \textbf{0.0828}          & \textbf{0.0477} & 0.0405                           & \textbf{0.0449} & \textbf{0.0337} & \textbf{0.0488} & 0.0541                        & \textbf{0.0291} & \textbf{0.0391} & \textbf{0.0467} \\ 
One & Ours       & 0.0872 & 0.0563 & \textbf{0.0400} & 0.0511 & 0.0379 & 0.0594 & \textbf{0.0376} & 0.0522 & 0.0463 & 0.0520       \\
\\
\multicolumn{12}{c}{Average precision at $15^{\circ}$ azimuth error ($\uparrow$)} \\
\cmidrule(r){5-7}
One & MobilePose \cite{hou2020mobilepose} & 0.4376 & 0.4111 & 0.4413 & 0.5293 & 0.8780 & 0.6195 & 0.0893 & 0.6052 & 0.3934 & 0.4894\\
Two & Two-stage  \cite{ahmadyan2021objectron}  & 0.8234 & 0.7222          & 0.8003                           & 0.8030 & \textbf{0.9404} & \textbf{0.8840} & 0.6444                        & \textbf{0.8561} & 0.5860           & 0.7844 \\ 
One & Ours       &\textbf{0.8622} & \textbf{0.7323} & \textbf{0.9561} & \textbf{0.8226} & 0.9361 & 0.8822 & \textbf{0.8945} & 0.7966 & \textbf{0.6757} & \textbf{0.8398}      \\
\\
\multicolumn{12}{c}{Average precision at $10^{\circ}$ elevation error ($\uparrow$)} \\
\cmidrule(r){5-7}
One & MobilePose \cite{hou2020mobilepose} & 0.7130 & 0.6289 & 0.6999 & 0.5233 & 0.8030 & 0.7053 & 0.6632 & 0.5413 & 0.4947 & 0.6414\\
Two & Two-stage  \cite{ahmadyan2021objectron}  & \textbf{0.9390} & \textbf{0.8616} & 0.8567                           & 0.8437 & \textbf{0.9476} & \textbf{0.9272} & 0.8365               & \textbf{0.7593} & 0.7544          & \textbf{0.8584} \\
One & Ours       & 0.9072 & 0.8535 & \textbf{0.8881} & \textbf{0.8704} & 0.9467 & 0.8999 & \textbf{0.8562} & 0.6922 & \textbf{0.7900} & 0.8560      \\
\bottomrule
\end{tabular}
\end{table*}

\begin{table*}[t]
\caption{Different strategies for 2D Keypoint output decoding  (average precision at 0.5 3D IoU metric ($\uparrow$)).
}
\centering
\begin{tabularx}{\textwidth}{cccccccccccc}
\toprule
     Strategy      & w/o add. proc. & Bike   & Book   & Bottle$^{*}$ & Camera & Cereal\_box & Chair  & Cup$^{*}$ & Laptop & Shoe   & Mean   \\
\midrule
Displacement & \textcolor{darkgreen}{\cmark}  & 0.6254 & 0.5263 & 0.7917 & \textbf{0.7191} & 0.8115 & \textbf{0.8492} & 0.7553 & 0.6737 & \textbf{0.6688} & 0.7134  \\      
Heatmap & \textcolor{darkgreen}{\cmark}  & 0.5788 & 0.5539 & 0.7970  & 0.7035 & 0.8138 & 0.8260  & 0.7626 & 0.6124 & 0.6079 & 0.6951  \\     
     
Distance~\cite{zhou2019objects} & \textcolor{darkred}{\xmark} & 0.6305 & 0.5436 & 0.7837 & 0.7111 & 0.8044 & 0.8460  & 0.7640  & 0.6692 & 0.6529 & 0.7117  \\
Sampling~\cite{lee2020guided}   &  \textcolor{darkred}{\xmark}                           & 0.6279 & 0.5516 & 0.7873 & 0.7182 & 0.8134 & 0.8466 & 0.7687 & 0.6751 & 0.6641 & 0.7170  \\
\midrule
{ Disp.~+ Heatmap}   &  \textcolor{darkgreen}{\cmark}      & \textbf{0.6419} & \textbf{0.5565} & \textbf{0.8021} & 0.7188 & \textbf{0.8211} & 0.8471 & \textbf{0.7704} & \textbf{0.6766} & 0.6618 & \textbf{0.7218}                  \\
\bottomrule
\end{tabularx}
\label{tab:ablation_representation}
\end{table*}

\begin{table*}[t]
\caption{Different strategies for computing cuboid dimensions.}
  \centering
  \begin{threeparttable}
\begin{tabular}{ccccccccc}
\toprule
\multirow{2}{*}{Method}                        & \multicolumn{4}{c}{Mean cuboid dimension error ($\downarrow$)}                        & \multicolumn{4}{c}{Average precision at 0.5 3D IoU ($\uparrow$)}                                 \\
\cmidrule(r){2-5} \cmidrule(r){6-9}
                                              & Book            & Laptop          & Others         & Mean            & Book            & Laptop          & Others         & Mean            \\
\midrule
Keypoint lifting~\cite{hou2020mobilepose} (no dim. pred.)  &-               & -               & -               & -               & 0.3999          & 0.5159          & 0.6540          & 0.6104         \\
Estimated dim. (w/o convGRU)            & 0.8474          & 0.9124          & \textbf{0.2434} & 0.3849          & 0.5401          & 0.6378          & \textbf{0.7528} & 0.7164       \\
Estimated dim. (w/ convGRU)                                   & \textbf{0.7440} & \textbf{0.6799} & 0.2475          & \textbf{0.3507} & \textbf{0.5565} & \textbf{0.6766} & 0.7519          & \textbf{0.7218} \\
Ground truth dim. (oracle)               & \textit{0}      & \textit{0}      & \textit{0}      & \textit{0}      & \textit{0.6955} & \textit{0.6942} & \textit{0.7907} & \textit{0.7694}          \\
\bottomrule
\end{tabular}
  \end{threeparttable}
\label{tab:ablation_scale}
\end{table*}

\begin{figure}[t]
  \centering
  \vspace*{0.06in}
  \begin{tikzpicture}[inner sep = 0pt, outer sep = 0pt]
    \node[anchor=south west] (fnC) at (0in,0in)
      {\includegraphics[height=1in,clip=true,trim=0.9in 2.3in 0.9in 1.5in]{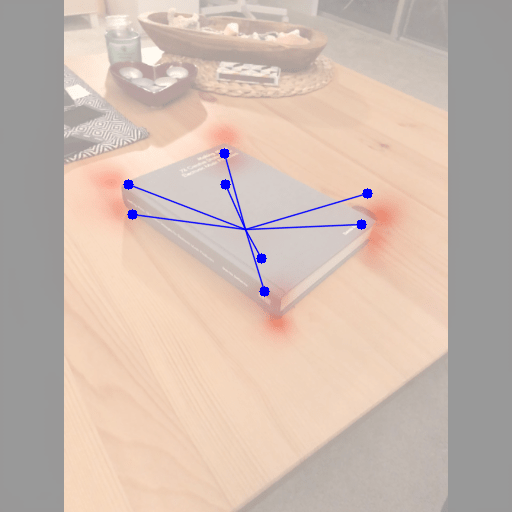}};
    \node[anchor=south west,xshift=2pt] (fnJ) at (fnC.south east)
      {\includegraphics[height=1in,clip=true,trim=0.9in 1.25in 0.9in 2.55in]{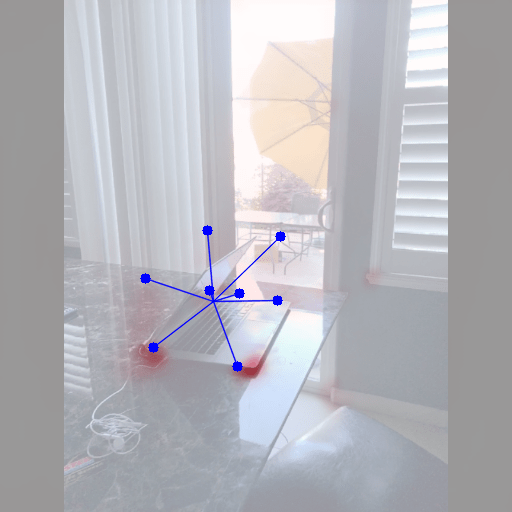}};
    \node[anchor=north west,xshift=2pt,yshift=1em] at (fnC.south west) 
      {\small Book (Cropped)};
    \node[anchor=north west,xshift=2pt,yshift=1em] at (fnJ.south west) 
      {\small Laptop (Cropped)};
  \end{tikzpicture}
  \caption{Two different keypoint representations.  
  Blue circles are found by the displacements, 
  while we overlay heatmap keypoints using red as intensity.
  Left: The heatmap is more accurate when the bounding box corners are visible and aligned with the object. 
  Right: Displacement performs better when the bounding box corners do not tightly fit the surface of the target 
  (such as the top of the laptop). \label{fig:exp_rep}} 
  \vspace*{-3 ex}
\end{figure}

\subsection{Different strategies for 2D Keypoint Output Decoding }

Most existing keypoint-based object pose estimation methods adopt either a heatmap \cite{oberweger2018making,tremblay2018deep} or displacement \cite{hu2019segmentation, hou2020mobilepose} representation for 2D keypoint detection. 
As shown in Figure~\ref{fig:exp_rep}, the large intra-class shape variance poses a key challenge for the keypoint representation.
Thus we designed an experiment to compare five different ways for post-processing the 2D keypoint output: 
1)~{\it Displacement} ignores the heatmap. 
2)~{\it Heatmap} ignores the displacement.
3)~{\it Distance} implements a heuristic similar to~\cite{zhou2019objects} that tries to select the more reliable point to use from the displacement or heatmap. 
4)~{\it Sampling}, inspired by \cite{lee2020guided}, fits a Gaussian mixture model to the heatmap peak estimate and the displacement prediction  for each keypoint and then samples $N$ points ($N=20$) to obtain a distribution of possible poses.
5)~{\it Our} proposed method keeps both displacement and heatmap.
As shown in Table~\ref{tab:ablation_representation}, our proposed combined method ({\it Displacement} + {\it Heatmap}) is better than either of the single representations ({\it Displacement} or {\it Heatmap}), and it also does not require additional processing (as in {\it Distance}~\cite{zhou2019objects} or {\it Sampling}~\cite{lee2020guided}). Thus, to balance accuracy and efficiency, we use this combined representation in all the other experiments.

\begin{figure}[t]
  \centering
  \vspace*{0.06in}
  \begin{tikzpicture}[inner sep = 0pt, outer sep = 0pt]
    \node[anchor=south west] (fnC) at (0in,0in)
      {\includegraphics[height=1.5in,clip=true,trim=0in 1.4in 0in 0.25in]{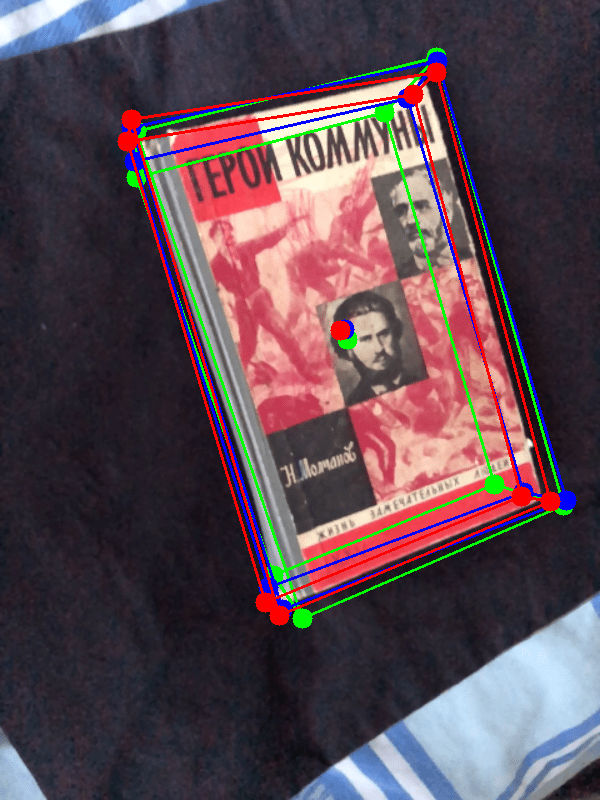}};
    \node[anchor=south west,xshift=2pt] (fnJ) at (fnC.south east)
      {\includegraphics[height=1.5in,clip=true,trim=0in 0in 0in 0.5in]{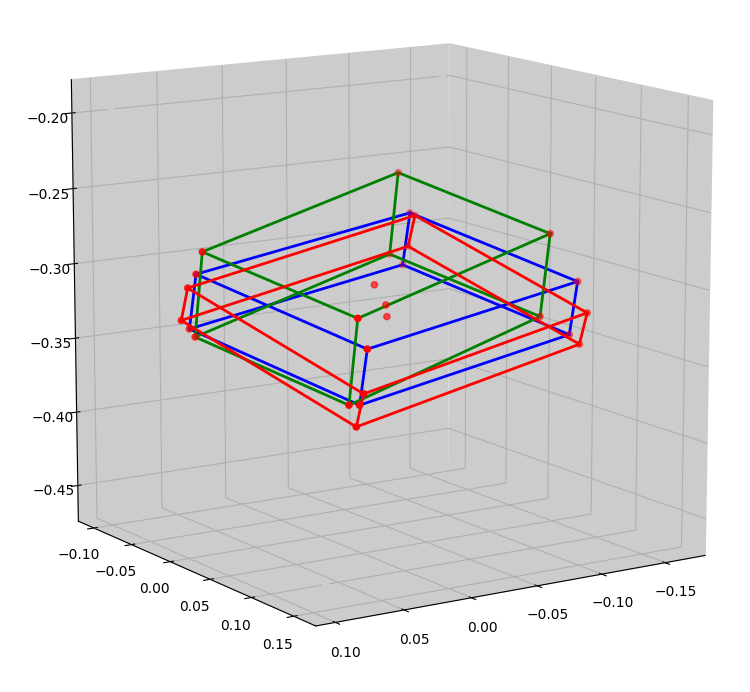}};
    \node[anchor=north west,xshift=2pt,yshift=1em] at (fnC.south west) 
      {\small \textcolor{white}{Book}};
  \end{tikzpicture}
  \caption{ Improvement due to the convGRU feature association module, where {\color{green}green} is ground truth, {\color{red}red} is our proposed method without feature association module, and {\color{blue}blue} is our proposed method with the module.
  When viewing a thin object from a certain perspective (azimuth close to $90^{\circ}$), 
  it is challenging to estimate the thickness of the target. 
  \label{fig:exp_scale}} 
\end{figure}

\subsection{Different Strategies for Cuboid Dimension Prediction}
In this section, we present an experiment on different strategies for cuboid dimension prediction, which further reveals the importance of accurate scale prediction and demonstrates the value of the sequential feature association module (convGRU) for hard cases.
We tested the following variants of our system: 
1) \textit{Keypoint lifting}, where we reimplemented the postprocessing part proposed by \cite{hou2020mobilepose} to retrieve the final pose using only the 2D projected cuboid keypoints;
2) \textit{No convGRU}, in which the convGRU layers were removed from our method (See Figure~\ref{fig:pipeline});
3) \textit{with convGRU}, our proposed method;   
4) \textit{Oracle}, which has access to the ground truth 3D aspect ratio (relative dimensions).
The results are shown in Table~\ref{tab:ablation_scale}, where we isolated two specific categories (book and laptop) since they have the greatest difference. The results indicate a strong relationship between 3D IoU result and the corresponding mean cuboid dimension error. For many categories (``Others" in Table~\ref{tab:ablation_scale}), the performance does not differ much, as well as their cuboid dimension prediction. Those instances are of similar aspect ratios and easier to estimate, \emph{e.g.}, bottles. 
On the other hand, the book and laptop categories are more challenging as the thickness of a book varies greatly while the laptop operates at different modes (whether the lid is open or closed).
The proposed convGRU module improves the prediction of their cuboid dimensions and leads to a better 3D IoU result.
Moreover, we found oracle with ground truth dimension achieved the best result while performance degraded using the simplified EP$n$P variant from \cite{hou2020mobilepose},
which suggests that predicting relative dimensions for category-level pose estimation from monocular RGB input is crucial to solving this problem.
Figure~\ref{fig:exp_scale} uses a particular example to show the ability of convGRU (blue) to retrieve the object 3D aspect ratio (relative dimensions)
when compared without convGRU (red). Even though both 2D keypoints look accurate, their scale predictions are different, leading to 3D IoU ($\uparrow$) improvement (0.5059 with convGRU {\em vs.} 0.3204  without convGRU).  

\subsection{Robot experiment}
\label{sec:robexp}

To demonstrate the potential of our object pose estimator on real-world applications, we investigated its ability for robotic manipulation.
We mounted a camera to the left wrist of a Baxter robot.\footnote{The camera is an Intel RealSense D415 depth camera, but we only used RGB images for this experiment.}
Previous work has explored different ways to obtain the scale factor, including manual measurement \cite{moons20093d, andrew2001multiple}, calculation based on normal vector of the table \cite{ahmadyan2021objectron}, pose fitting via depth alignment \cite{wang2019normalized}, and multi-view consistency \cite{lourakis2013accurate}. In this experiment, we manually measured the height of each object for simplicity.

We placed one shoe on the table, and another shoe in the right robot gripper.  The robot was then instructed to place the shoe in its gripper next to and aligned with the shoe on the table, using the position and orientation estimated by our proposed system.  We observed fairly reliable behavior by the robot on this task, with 4 out of 5 trials successful over a variety of previously unseen shoes. 
When comparing against prior work, such as~\cite{manuelli2019kpam} and~\cite{xu2021affordance}, 
our 3D oriented bounding box offers an alternative choice to the semantic 3D keypoint representation.  
Nevertheless, estimating scale reliably remains an unsolved problem, which we leave for future work.

\section{Conclusion}
We have presented a single-stage method for category-level 6-DoF pose prediction of previously unseen object instances from RGB input. 
Unlike many previous approaches, CAD models of instances are not needed at training nor test time, and complex annotations are not required for training.
For accurate 2D keypoint detection, we adopt a combined representation of both displacements and heatmaps to mitigate uncertainty. 
We also show the importance of precise cuboid dimension prediction for category-level pose estimation problem, and propose to use convGRU sequential feature association to further improve accuracy for challenging cases with varied aspect ratios.
Those design choices enable us to explore the potential of simple 3D bounding box annotations from a large-scale real-world dataset, without extra inputs like depth.
We demonstrate state-of-the-art performance on the large-scale real-world Objectron dataset, along with a robotic experiment indicating the potential of our proposed method to serve real-world applications. 
Future work will aim to improve the results by 
incorporating shape geometry embeddings, exploring lightweight backbone networks, and leveraging iterative post refinement.

\bibliographystyle{IEEEtran}
\bibliography{main.bib}

\begin{thebibliography}{10}
\providecommand{\url}[1]{#1}
\csname url@samestyle\endcsname
\providecommand{\newblock}{\relax}
\providecommand{\bibinfo}[2]{#2}
\providecommand{\BIBentrySTDinterwordspacing}{\spaceskip=0pt\relax}
\providecommand{\BIBentryALTinterwordstretchfactor}{4}
\providecommand{\BIBentryALTinterwordspacing}{\spaceskip=\fontdimen2\font plus
\BIBentryALTinterwordstretchfactor\fontdimen3\font minus
  \fontdimen4\font\relax}
\providecommand{\BIBforeignlanguage}[2]{{%
\expandafter\ifx\csname l@#1\endcsname\relax
\typeout{** WARNING: IEEEtran.bst: No hyphenation pattern has been}%
\typeout{** loaded for the language `#1'. Using the pattern for}%
\typeout{** the default language instead.}%
\else
\language=\csname l@#1\endcsname
\fi
#2}}
\providecommand{\BIBdecl}{\relax}
\BIBdecl

\bibitem{xiang2018posecnn}
Y.~Xiang, T.~Schmidt, V.~Narayanan, and D.~Fox, ``{PoseCNN}: A convolutional
  neural network for {6D} object pose estimation in cluttered scenes,'' in
  \emph{RSS}, 2018.

\bibitem{tremblay2018deep}
J.~Tremblay, T.~To, B.~Sundaralingam, Y.~Xiang, D.~Fox, and S.~Birchfield,
  ``Deep object pose estimation for semantic robotic grasping of household
  objects,'' in \emph{CoRL}, 2018, pp. 306--316.

\bibitem{wang2019densefusion}
C.~Wang, D.~Xu, Y.~Zhu, R.~Mart{\'\i}n-Mart{\'\i}n, C.~Lu, L.~Fei-Fei, and
  S.~Savarese, ``{DenseFusion}: {6D} object pose estimation by iterative dense
  fusion,'' in \emph{CVPR}, 2019, pp. 3343--3352.

\bibitem{calli2015ram:ycb}
B.~Calli, A.~Walsman, A.~Singh, S.~Srinivasa, P.~Abbeel, and A.~M. Dollar,
  ``Benchmarking in manipulation research: {U}sing the {Yale-CMU-Berkeley}
  object and model set,'' \emph{IEEE Robotics and Automation Magazine},
  vol.~22, no.~3, Sep. 2015.

\bibitem{pavlakos2017}
G.~Pavlakos, X.~Zhou, A.~Chan, K.~G. Derpanis, and K.~Daniilidis, ``{6-DoF}
  object pose from semantic keypoints,'' in \emph{ICRA}, 2017, pp. 2011--2018.

\bibitem{wang2019normalized}
H.~Wang, S.~Sridhar, J.~Huang, J.~Valentin, S.~Song, and L.~J. Guibas,
  ``Normalized object coordinate space for category-level {6D} object pose and
  size estimation,'' in \emph{CVPR}, 2019, pp. 2642--2651.

\bibitem{ke2020gsnet}
L.~Ke, S.~Li, Y.~Sun, Y.-W. Tai, and C.-K. Tang, ``{GSNet}: Joint vehicle pose
  and shape reconstruction with geometrical and scene-aware supervision,'' in
  \emph{ECCV}, 2020, pp. 515--532.

\bibitem{tian2020shape}
M.~Tian, M.~H. Ang, and G.~H. Lee, ``Shape prior deformation for categorical
  {6D} object pose and size estimation,'' in \emph{ECCV}, 2020, pp. 530--546.

\bibitem{chen2020learning}
D.~Chen, J.~Li, Z.~Wang, and K.~Xu, ``Learning canonical shape space for
  category-level {6D} object pose and size estimation,'' in \emph{CVPR}, 2020,
  pp. 11\,973--11\,982.

\bibitem{chen2020category}
X.~Chen, Z.~Dong, J.~Song, A.~Geiger, and O.~Hilliges, ``Category level object
  pose estimation via neural analysis-by-synthesis,'' in \emph{ECCV}, 2020, pp.
  139--156.

\bibitem{manhardt2020cps++}
F.~Manhardt, G.~Wang, B.~Busam, M.~Nickel, S.~Meier, L.~Minciullo, X.~Ji, and
  N.~Navab, ``{CPS++}: Improving class-level {6D} pose and shape estimation
  from monocular images with self-supervised learning,'' \emph{arXiv preprint
  arXiv:2003.05848}, 2020.

\bibitem{shapenet2015}
A.~X. Chang, T.~Funkhouser, L.~Guibas, P.~Hanrahan, Q.~Huang, Z.~Li,
  S.~Savarese, M.~Savva, S.~Song, H.~Su, J.~Xiao, L.~Yi, and F.~Yu,
  ``{{ShapeNet}: An Information-Rich {3D} Model Repository},'' \emph{arXiv
  preprint arXiv:1512.03012}, 2015.

\bibitem{kortylewski2018training}
A.~Kortylewski, A.~Schneider, T.~Gerig, B.~Egger, A.~Morel-Forster, and
  T.~Vetter, ``Training deep face recognition systems with synthetic data,''
  \emph{arXiv preprint arXiv:1802.05891}, 2018.

\bibitem{hou2020mobilepose}
T.~Hou, A.~Ahmadyan, L.~Zhang, J.~Wei, and M.~Grundmann, ``{MobilePose}:
  Real-time pose estimation for unseen objects with weak shape supervision,''
  \emph{arXiv preprint arXiv:2003.03522}, 2020.

\bibitem{ahmadyan2021objectron}
A.~Ahmadyan, L.~Zhang, A.~Ablavatski, J.~Wei, and M.~Grundmann, ``Objectron: A
  large scale dataset of object-centric videos in the wild with pose
  annotations,'' in \emph{CVPR}, 2021, pp. 7822--7831.

\bibitem{zhou2019objects}
X.~Zhou, D.~Wang, and P.~Kr{\"a}henb{\"u}hl, ``Objects as points,'' \emph{arXiv
  preprint arXiv:1904.07850}, 2019.

\bibitem{ballas2016delving}
N.~Ballas, L.~Yao, C.~Pal, and A.~C. Courville, ``Delving deeper into
  convolutional networks for learning video representations.'' in \emph{ICLR},
  2016.

\bibitem{gao2020monocular}
T.~Gao, H.~Pan, and H.~Gao, ``Monocular {3D} object detection with sequential
  feature association and depth hint augmentation,'' \emph{arXiv preprint
  arXiv:2011.14589}, 2020.

\bibitem{zeng2017multi}
A.~Zeng, K.-T. Yu, S.~Song, D.~Suo, E.~Walker, A.~Rodriguez, and J.~Xiao,
  ``Multi-view self-supervised deep learning for {6D} pose estimation in the
  amazon picking challenge,'' in \emph{ICRA}, 2017, pp. 1386--1383.

\bibitem{li2018deepim}
Y.~Li, G.~Wang, X.~Ji, Y.~Xiang, and D.~Fox, ``{DeepIM}: Deep iterative
  matching for {6D} pose estimation,'' in \emph{ECCV}, 2018, pp. 683--698.

\bibitem{sundermeyer2018implicit}
M.~Sundermeyer, Z.-C. Marton, M.~Durner, M.~Brucker, and R.~Triebel, ``Implicit
  {3D} orientation learning for {6D} object detection from {RGB} images,'' in
  \emph{ECCV}, 2018, pp. 699--715.

\bibitem{choi20123d}
C.~Choi and H.~I. Christensen, ``{3D} textureless object detection and
  tracking: An edge-based approach,'' in \emph{IROS}, 2012, pp. 3877--3884.

\bibitem{birdal2015point}
T.~Birdal and S.~Ilic, ``Point pair features based object detection and pose
  estimation revisited,'' in \emph{3DV}, 2015, pp. 527--535.

\bibitem{hodan2018bop}
T.~Hoda{\v{n}}, F.~Michel, E.~Brachmann, W.~Kehl, A.~Glent~Buch, D.~Kraft,
  B.~Drost, J.~Vidal, S.~Ihrke, X.~Zabulis, C.~Sahin, F.~Manhardt, F.~Tombari,
  T.-K. Kim, J.~Matas, and C.~Rother, ``{BOP}: Benchmark for {6D} object pose
  estimation,'' \emph{ECCV}, 2018.

\bibitem{rad2017bb8}
M.~Rad and V.~Lepetit, ``{BB8}: A scalable, accurate, robust to partial
  occlusion method for predicting the {3D} poses of challenging objects without
  using depth,'' in \emph{ICCV}, 2017, pp. 3828--3836.

\bibitem{tekin2018real}
B.~Tekin, S.~N. Sinha, and P.~Fua, ``Real-time seamless single shot {6D} object
  pose prediction,'' in \emph{CVPR}, 2018, pp. 292--301.

\bibitem{oberweger2018making}
M.~Oberweger, M.~Rad, and V.~Lepetit, ``Making deep heatmaps robust to partial
  occlusions for {3D} object pose estimation,'' in \emph{ECCV}, 2018, pp.
  119--134.

\bibitem{peng2019pvnet}
S.~Peng, Y.~Liu, Q.~Huang, X.~Zhou, and H.~Bao, ``{PVNet}: Pixel-wise voting
  network for 6{DoF} pose estimation,'' in \emph{CVPR}, 2019, pp. 4561--4570.

\bibitem{song2020hybridpose}
C.~Song, J.~Song, and Q.~Huang, ``{HybridPose}: {6D} object pose estimation
  under hybrid representations,'' in \emph{CVPR}, 2020, pp. 431--440.

\bibitem{he2017mask}
K.~He, G.~Gkioxari, P.~Doll{\'a}r, and R.~Girshick, ``Mask {R-CNN},'' in
  \emph{ICCV}, 2017, pp. 2961--2969.

\bibitem{lepetit2009epnp}
V.~Lepetit, F.~Moreno-Noguer, and P.~Fua, ``{EP$n$P}: An accurate {$O(n)$}
  solution to the {P$n$P} problem,'' \emph{IJCV}, vol.~81, no.~2, p. 155, 2009.

\bibitem{liu2020smoke}
Z.~Liu, Z.~Wu, and R.~T{\'o}th, ``{SMOKE}: Single-stage monocular {3D} object
  detection via keypoint estimation,'' in \emph{CVPR Workshops}, 2020, pp.
  996--997.

\bibitem{wang2020centermask}
Y.~Wang, Z.~Xu, H.~Shen, B.~Cheng, and L.~Yang, ``{CenterMask}: {Single} shot
  instance segmentation with point representation,'' in \emph{CVPR}, 2020, pp.
  9313--9321.

\bibitem{yu2018deep}
F.~Yu, D.~Wang, E.~Shelhamer, and T.~Darrell, ``Deep layer aggregation,'' in
  \emph{CVPR}, 2018, pp. 2403--2412.

\bibitem{zhu2019deformable}
X.~Zhu, H.~Hu, S.~Lin, and J.~Dai, ``Deformable {Convnets} v2: More deformable,
  better results,'' in \emph{CVPR}, 2019, pp. 9308--9316.

\bibitem{abdel2015direct}
Y.~I. Abdel-Aziz, H.~M. Karara, and M.~Hauck, ``Direct linear transformation
  from comparator coordinates into object space coordinates in close-range
  photogrammetry,'' \emph{Photogrammetric Engineering \& Remote Sensing},
  vol.~81, no.~2, pp. 103--107, 2015.

\bibitem{lin2017focal}
T.-Y. Lin, P.~Goyal, R.~Girshick, K.~He, and P.~Doll{\'a}r, ``Focal loss for
  dense object detection,'' in \emph{ICCV}, 2017, pp. 2980--2988.

\bibitem{lee2020guided}
M.~A. Lee, C.~Florensa, J.~Tremblay, N.~Ratliff, A.~Garg, F.~Ramos, and D.~Fox,
  ``{Guided Uncertainty-Aware Policy Optimization}: Combining learning and
  model-based strategies for sample-efficient policy learning,'' in
  \emph{ICRA}, 2020, pp. 7505--7512.

\bibitem{hu2019segmentation}
Y.~Hu, J.~Hugonot, P.~Fua, and M.~Salzmann, ``Segmentation-driven {6D} object
  pose estimation,'' in \emph{CVPR}, 2019, pp. 3385--3394.

\bibitem{moons20093d}
T.~Moons, L.~Van~Gool, and M.~Vergauwen, \emph{{3D} reconstruction from
  multiple images: Principles}.\hskip 1em plus 0.5em minus 0.4em\relax Now
  Foundations and Trends, 2009.

\bibitem{andrew2001multiple}
R.~Hartley and A.~Zisserman, \emph{Multiple view geometry in computer
  vision}.\hskip 1em plus 0.5em minus 0.4em\relax Cambridge Univ. Press, 2001.

\bibitem{lourakis2013accurate}
M.~Lourakis and X.~Zabulis, ``Accurate scale factor estimation in {3D}
  reconstruction,'' in \emph{International Conference on Computer Analysis of
  Images and Patterns}, 2013, pp. 498--506.

\bibitem{manuelli2019kpam}
L.~Manuelli, W.~Gao, P.~Florence, and R.~Tedrake, ``{kPAM}: Keypoint
  affordances for category-level robotic manipulation,'' \emph{ISRR}, 2019.

\bibitem{xu2021affordance}
R.~Xu, F.-J. Chu, C.~Tang, W.~Liu, and P.~A. Vela, ``An affordance keypoint
  detection network for robot manipulation,'' \emph{RA-L}, vol.~6, no.~2, pp.
  2870--2877, 2021.

\end{thebibliography}

\end{document}